\documentclass[letterpaper, 10 pt, conference]{ieeeconf}  

\usepackage{amsmath} 
\usepackage{amssymb}  
\usepackage{graphicx}
\usepackage{url}
\DeclareUnicodeCharacter{03A7}{\ensuremath{\chi}}
\usepackage{hyperref}
\usepackage{graphicx}
\usepackage{framed,multirow}
\usepackage{makecell}
\usepackage{xcolor}
\usepackage{booktabs}

\title{\LARGE \bf
GazeMoE: Perception of Gaze Target with Mixture-of-Experts}

\author{Zhuangzhuang Dai$^{*,1}$, Zhongxi Lu$^{2}$, Vincent G. Zakka$^{1}$, Luis J. Manso$^{1}$, Jose M Alcaraz Calero$^{1}$, Chen Li$^{3}$\\
\normalsize $^{1}$Dept. of Applied AI and Robotics,
         Aston University, Birmingham, United Kingdom\\
\normalsize $^{2}$Computing Science,
         University of Leicester, Leicester, United Kingdom\\
\normalsize $^{3}$Dept. of Materials and Production, Aalborg University, Aalborg, Denmark\\
{$^{*}$Corresponding address: \tt\small z.dai1@aston.ac.uk}
}

\begin{document}

\maketitle
\thispagestyle{empty}
\pagestyle{empty}

\begin{abstract}
Estimating human gaze target from visible images is a critical task for robots to understand human attention, yet the development of generalizable neural architectures and training paradigms remains challenging. While recent advances in pre-trained vision foundation models offer promising avenues for locating gaze targets, the integration of multi-modal cues— including eyes, head poses, gestures, and contextual features—demands adaptive and efficient decoding mechanisms. Inspired by Mixture-of-Experts (MoE) for adaptive domain expertise in large vision-language models, we propose GazeMoE, a novel end-to-end framework that selectively leverages gaze-target-related cues from a frozen foundation model through MoE modules. To address class imbalance in gaze target classification (in-frame vs. out-of-frame) and enhance robustness, GazeMoE incorporates a class-balancing auxiliary loss alongside strategic data augmentations, including region-specific cropping and photometric transformations. Extensive experiments on benchmark datasets demonstrate that our GazeMoE achieves state-of-the-art performance, outperforming existing methods on challenging gaze estimation tasks. The code and pre-trained models are released at: \url{https://huggingface.co/zdai257/GazeMoE}.
\end{abstract}

\section{Introduction}

Gaze serves as a fundamental indicator of human intention, interest, and cognitive state, making its accurate estimation critical for a wide range of applications. Gaze plays a forefront role for robots and autonomous systems to understand user engagement~\cite{engagechild2020}, driver fatigue~\cite{driverfatiguereview2022}, consumer interest~\cite{goo2021}, and autism through identifying abnormal visual attention patterns~\cite{childplay2023}. Estimation of gaze targets has gained increasing traction in the past decade~\cite{Hu2022Gaze, Liu2019A}. Unlike gaze direction analysis, gaze target aims to derive where a person is looking through absorbing the semantics of a scene. 

\begin{figure}[t]
    \centering
    \includegraphics[width=0.99\linewidth]{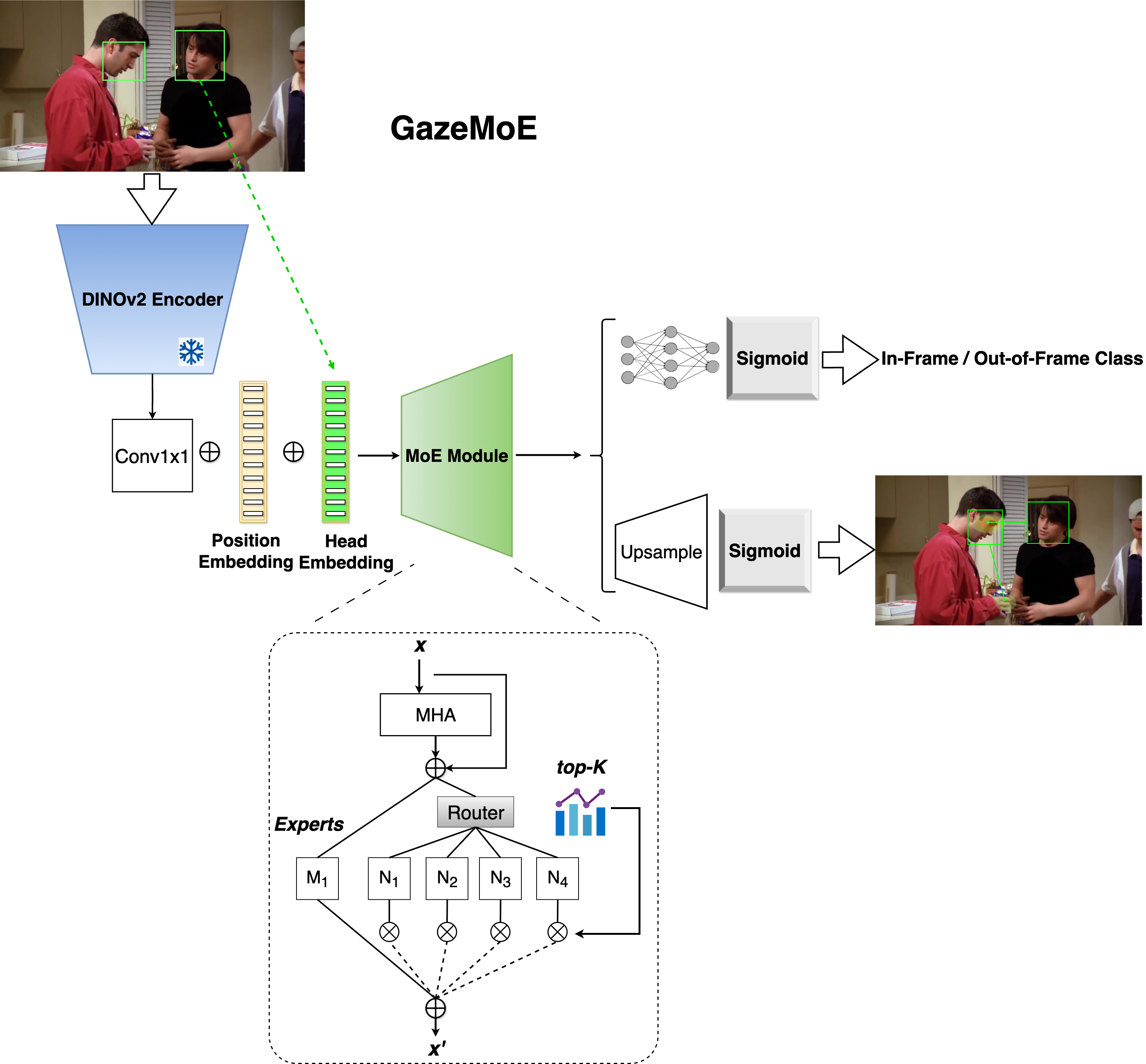}
    \caption{GazeMoE data flow diagram. A frozen DINOv2 is used to extract fine-grained scene representations alongside the Mixture-of-Experts (MoE) module specialized in selectively routing gaze-related cues.}
    \label{fig:flow}
\end{figure}

Directly estimating subjects' attended targets from visible data unlocks more advanced applications such as joint attention detection~\cite{chong2018}, human-robot interaction~\cite{dai_icac2023}, or intention prediction~\cite{gazedriverintent2019}, and provides the additional benefit of being less invasive than other alternatives such as headsets. GazeFollow~\cite{gazefollow2015} marks the first sizeable dataset in the community. By leveraging standard cameras and Deep Neural Networks, data-driven methods have grown rapidly since then. However, successes are limited by specific gaze evaluation setting, such as movies~\cite{attendedtarget2020} or early education~\cite{childplay2023}. It is yet difficult to use such models to predict human gaze targets reliably in real-world scenarios. In the literature, gaze target has been found highly dependent on four visual cues: (a) visible eye landmarks~\cite{deepeyecontact2020,gazefollow2015}, (b) head pose~\cite{gupta2022a}, (c) gestures and geometry of the scene~\cite{3dgaze2018,gazefollow3d2023}, and (d) contextual saliency~\cite{gazefollow2015,attendedtarget2020}. In reality, gaze targets can be more diverse than those annotated in datasets, which tend to curate data from lifelogging, Youtube, and other public imaging databases. Moreover, out-of-frame gaze targets, in which a subject fixates at things outside camera's view, will be more prevalent and unpredictable than in-frame perceivable objects in real-world scenarios~\cite{gt360}.

Recently, vision foundation models demonstrated outstanding performance~\cite{gazelle2024,vitgaze2024} proving that representations of eyes, head poses, gestures, and environmental saliency have been superbly learned in a DINOv2 model~\cite{dinov2}. Unfortunately, prior works have not thoroughly investigated an optimized neural architecture of decoder to selectively consume scene-specific cues for effective gaze target evaluation, nor the optimal augmentation and training strategies. How to allow models to adapt to missing features, e.g., due to occlusions or image distortion, remains largely unexplored.

The Mixture-of-Experts (MoE)~\cite{Masoudnia2012Mixture, moe_theory2018} is a neural network block designed to improve model capacity and efficiency by dynamically routing input data to specialized sub-networks, i.e., experts~\cite{moe_survey}. Instead of processing all inputs through the entire network, MoE selectively activates only a subset of experts for each input, enabling scalable and computationally efficient learning. We note that gaze target estimation requires adaptively integrating eyes, head, gesture, and contextual information~\cite{attendedtarget2020}, some of which may or may not be available in different scenes. In this work, we propose an end-to-end model, GazeMoE, leveraging a shared expert and four routed experts that selectively gate input features via ranking independent feedforward networks in transformer decoders as shown in Fig.~\ref{fig:flow}. Each expert learns distinct features which improves representation diversity and model performance on different gaze data distributions. GazeMoE leverages an encoder-decoder architecture~\cite{gazelle2024,vitgaze2024} with a frozen DINOv2 of ViT$_{large}$~\cite{dinov2} as encoder to extract low-level, gaze-related spatial representation. Two output heads are designed to predict gaze target location as a heatmap of probability and to classify in-frame or out-of-frame cases, respectively. The modular structure of GazeMoE helps avoid overfitting by distributing learning across experts. Since only a few experts are used per input, GazeMoE preserves computational cost while achieving new state-of-the-art performance on various datasets.

Previous methods predominantly use a Binary Cross-Entropy (BCE) loss for in-frame or out-of-frame classification, overlooking class imbalance in common gaze datasets. In GazeMoE, a focal loss is adopted to address class imbalance by focusing more on hard-to-classify minority samples. To maximize the performance as well as generalization, GazeMoE encompasses a full stack of augmentations, including random cropping, horizontal flipping, head prompt jittering, and photometric changes (colour jittering, random grayscaling, sharpness and contrast). It has been found that such augmentation techniques are essential to optimize training. GazeMoE achieves state-of-the-art performance on all common benchmarks. Importantly, it shows excellent robustness and generalization in out-of-distribution visual scenes, such as fisheye lens imaging~\cite{gazefollowing360_2021} and children gaze estimation~\cite{childplay2023}.

The main contributions of this work are as follows. (1) We propose GazeMoE, an end-to-end gaze target estimator leveraging Mixture-of-Experts modules that produces state-of-the-art results on a variety of benchmarks; (2) We show optimal training strategies including the introduction of a custom class-balancing loss and a suite of augmentations; (3) Extensive experiments demonstrate the superiority of our proposed GazeMoE in accuracy, robustness, and generalization. Particularly, GazeMoE is the first-ever attempt that demonstrates reliable gaze target evaluation across both fisheye lens imaging and standard lens imaging domains.

\section{Related Work}
\label{sec:relatedwork}

\textbf{Data-driven Gaze Target Evaluation}. Ever since Recasens \textit{et al.}~\cite{gazefollow2015} curated the GazeFollow challenge -- a large scale dataset aimed at locating subjects' gaze target in 2D images, researchers have been pushing the boundaries of gaze target estimation in terms of precision~\cite{chong2018,attendedtarget2020,objectaware_gt2023,vitgaze2024,gazelle2024}, generalization~\cite{goo2021,childplay2023}, efficiency~\cite{attendedtarget2020,gazelle2024}, as well as 3D scene semantics~\cite{Fang_dual_attention2021,escnet2022,objectaware_gt2023}. Traditional methods typically focus on fusing human facial and gestural features with scene context through multiple encoder pathways~\cite{attendedtarget2020,Fang_dual_attention2021} or with the aid of multi-channel sensing modalities~\cite{rgbd_gaze2022}. They realize successes in common benchmarks constructed for specific scenarios, such as TV shows~\cite{chong2018} or kids behaviour monitoring~\cite{childplay2023}, whereas, engendering trade-offs in generalization, complexity of multi-objective training, and runtime efficiency.

Ryan~\cite{gazelle2024} and Song~\cite{vitgaze2024} recently showed that a large pre-trained vision foundation model, like DINOv2~\cite{dinov2}, is effective at encoding gaze target-related information. A frozen pre-trained encoder combined with a decoder deliver state-of-the-art performance across many benchmarks and generalized well for out-of-distribution data. In Gaze-LLE~\cite{gazelle2024}, head positions are incorporated as positional embeddings without the need for a separate learnable pathway. A compact, transformer-based decoder can jointly locate gaze targets and classify whether a target is in or outside a frame~\cite{attendedtarget2020}. GazeTarget360~\cite{gt360} is recently proposed -- a system leveraging multi-scale fusion in decoder -- for efficient gaze target evaluation incorporating eye contact detection. However, the gaze target localization precision is suboptimal.

\vspace{0.2cm}
\noindent
\textbf{Training Paradigm}. Different learning targets have been explored to train a gaze target estimator. Using L2 loss to supervise predicted heatmaps is adopted by most works~\cite{gazefollow2015,chong2018,Fang_dual_attention2021,objectaware_gt2023,vitgaze2024} in the community. Some recent works~\cite{gazefollowing360_2021,gazelle2024} found pixel-wise Binary Cross-Entropy loss more effective in deriving fine-grained gaze target. Cosine distance of predicted gaze vectors or head poses are utilized as auxiliary losses which prove effective to guide the learning process for many~\cite{chong2018,childplay2023,objectaware_gt2023,vitgaze2024}.

What remains to be answered is what the optimal decoder network is and its training paradigm. Recently, Large-Language Models benefit tremendously from MoE architectures which allow dynamic input routing~\cite{moe_survey}. Would a MoE-based decoder bring more robustness to gaze target estimation which depends on selective scene features? What is the best loss function? Data augmentation plays a crucial role in computer vision but none of the prior work has concurred in the best augmentations for gaze target tasks. In this work, we are motivated to address these questions.

\section{Method}
\label{sec:method}

Given an input image $I \in \mathbb{R}^{h\times w\times 3}$, gaze target estimation aims to localize the gaze attended position in pixel coordinates, and discern whether the target is inside or outside the camera's field of view. Suppose there are multiple subjects in the scene, their head bounding boxes, $bbox_i = [x_{min}, y_{min}, x_{max}, y_{max}]_i$, form the other input pathway of head prompts. The gaze target task can be expressed as

\begin{equation}
    y, hm = F_{\theta}(I, bbox)
\end{equation}

\noindent
where $y$ is the probability of an in-frame target; $hm$ is a heatmap containing target location probabilities within the frame $I$; and $\theta(\cdot)$ represents all parameters of a neural network.

\subsection{Mixture-of-Expert Module}

\begin{figure*}[!t]
    \centering
    \includegraphics[width=1.0\linewidth]{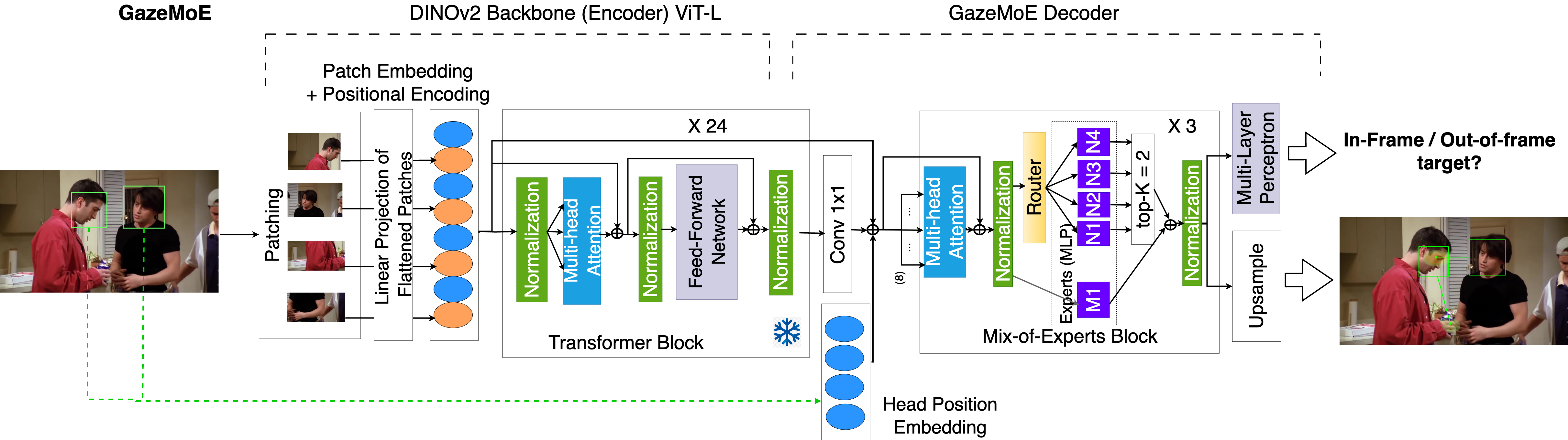}
    \caption{GazeMoE architecture. Given an input image, the model predicts whether a person's gaze target is in-frame or out-of-frame and where target is. A frozen DINOv2 ViT-L backbone is used to extract fine-grained scene representations. The GazeMoE decoder with three Mix-of-Experts blocks is specialized in selectively routing gaze-target-related features.}
    \label{fig:architecture}
\end{figure*}

We notice visual scenes of gaze target tasks vary enormously necessitating versatile representation learning of the scene depending on availability of different visual cues. Previous methods predominately exploit handcrafted cues including eye appearance~\cite{vitgaze2024}, head poses~\cite{gupta2022a}, or a conical field-of-view from a head adhering to straight-line propagation of light~\cite{Fang_dual_attention2021}. However, eyes and head poses may be occluded when subjects look away from camera~\cite{gazefollow2015} imposing high uncertainties. The rule-of-thumb that eyes and gaze target should be line-of-sight connected no longer holds true under imaging distortion, especially in panorama images where sphere-to-plane projection brings severe distortion~\cite{gazefollowing360_2021}. Moreover, traditional methods often find performance degradation in different subject groups like children or autism patients~\cite{childplay2023}. Recent work has shown that a separate encoder pathway for head features is unnecessary, in which head positional embeddings suffice~\cite{gazelle2024}. A convolutional layer with $1\times1$ kernel can be used to pull features from a frozen DINOv2 encoder which gives robust representations of human gaze~\cite{vitgaze2024}. Subsequently, these latent features can be decoded with standard transformer blocks.

To this end, we propose an MoE module to enhance feature feed-forward propagation in transformer decoder blocks, as illustrated in Figure~\ref{fig:architecture}. The MoE module can adaptively integrate shared and scenario-specific (selectively routed) knowledge based on the input image~\cite{moe_survey,moe2024}. This allows for segregation of routed information, i.e., the four pivotal gaze target cues -- eye appearance, head poses, gestures, and contextual saliency, when some are available and others are not across different scenes. MoE acts as a strong and efficient gated decoder to select only relevant cues as contender for gaze target evaluation whilst pruning visual noises. Mixing shared experts and routed experts enable pathways for both common knowledge and conditional features which stabilize model learning.

Given the self-attention scores acquired by Multi-Head Attention (MHA) in each transformer block, input to the feed-forward network (FFN), $x \in \mathbb{R}^{B \times Seq \times d_{model}}$, will be flattened before the MoE module, where $B$ is batch size and $Seq$ is sequence length. Each shared expert, $E_j^{shared}$, will contribute to shared features,

\begin{equation}
       x_{shared} = \frac{1}{M} \sum_{j=1}^M E_j^{shared}(x) 
\end{equation}
\begin{equation}
       E_j^{shared}(x) = GeLU(W_{j,1} \cdot x) W_{j,2}
\end{equation}

\noindent
where $E_j^{shared}$ are MLP with weights $W_{j,1} \in \mathbb{R}^{d_{model} \times d_h}$ and $W_{j,2} \in \mathbb{R}^{d_h \times d_{model}}$ with a GeLU activation in between; $M$ is the number of shared experts; $d_h$ represents hidden layer dimension of the shared expert.

The $N$ routed experts, $E_i^{routed}$, take an identical shape as each shared expert. Routed experts will go through a gating mechanism where the gating network $W_g \in \mathbb{R}^{d_{model} \times N}$ will select top-$K$ routed experts to use for each input. The gating weights can be expressed as,

         \begin{equation}
             g(x) = \sigma(W_g \cdot x) \quad \in \mathbb{R}^N
         \end{equation}

\noindent
where $\sigma$ is Softmax function. The top-$K$ selection reads;

         \begin{align}
             \text{Top-}K \text{ indices: } &\{i_1, ..., i_K\} = \text{argsort}(g(x))[:K] \\
             \text{Normalized weights: } &w_k = \frac{g(x)_{i_k}}{\sum_{k'=1}^K g(x)_{i_{k'}}}
         \end{align}
    
The routed feature is comprised of weighted sum of routed experts' outputs;

         \begin{equation}
             x_{routed} = \sum_{k=1}^K w_k \cdot E_{i_k}^{routed}(x)
        \end{equation}

Final output of the MoE module can be written as;

   \begin{equation}
       x' = x_{shared} + x_{routed}
   \end{equation}

In this work, we set $N=4$ routed experts to selectively capture the four gaze target cues identified in previous work; $M=1$ shared expert for passage of common scene features; each expert with $d_{model}=256$ hidden size in MLP layers; and $K=2$ meaning only the top-$2$ weighted experts will be used in the calculations for optimized performance and efficiency. Detailed validation of these parameters is discussed later in the Ablation Study section.

\subsection{Heatmap Loss and Auxiliary Focal Loss}

Previous works mostly adopted heatmap L2, i.e., Mean-Squared Error (MSE) loss~\cite{Dokania_2024_BMVC,chong2018,escnet2022}, sometimes combined with gaze angular loss~\cite{Fang_dual_attention2021} for training. Other research reveal pixel-wise BCE loss produces satisfactory performance~\cite{gazelle2024,gazefollowing360_2021} raising concerns about an optimal loss setting. The BCE loss treats each pixel as an independent binary classification. This aligns well with the heatmap's role of modelling probability distribution across the gaze target region. Our experiments show strong evidence that pixel-wise BCE loss is less sensitive to the magnitude of probabilistic error than MSE which may overly penalize peak (0.9 prediction vs. 1.0 ground-truth gaze) or valleys (0.1 prediction vs. 0 ground-truth gaze). Given gaze target is taken as the max probability coordinate, the absolute probability values do not matter as much. Thus, BCE loss generally handles noisy or imperfect target heatmaps better.

As for in/out classification, the vanilla BCE loss can be expressed as;

\begin{equation}
\text{BCE}(p, y) = -y \log(p) - (1 - y) \log(1 - p)
\end{equation}

\noindent
where $p$ denotes probability of an in-frame target and $y$ is the ground-truth label. We introduce a focal loss component to allow better robustness for in/out-of-frame classification tasks of imbalanced classes, which is typical in related datasets. For instance, VideoAttentionTarget~\cite{attendedtarget2020} contains approximately 40\% out-of-frame samples, while ChildPlay~\cite{childplay2023} has fewer than 10\% out-of-frame samples. To address class imbalance, let

\begin{equation}
p_t = 
\begin{cases}
p & \text{if } y = 1 \\
1 - p & \text{if } y = 0
\end{cases}
\end{equation}

\noindent
then, the focal loss can be written as

\begin{equation}
\mathcal{L}_{focal} = \alpha \cdot (1 - p_t)^{\gamma} \cdot \text{BCE}(p, y)
\end{equation}

\noindent
where $\alpha$ is the class weighting factor to handle class imbalance and $\gamma$ is the focusing parameter. We set $\alpha$ as the ratio between minor class and major class in our experiments and $\gamma$ is fixed at $2.0$. This allows enhanced learning of minority classes and difficult samples.

The final loss function is expressed as,

\begin{equation}
\label{eqn:loss}
    \mathcal{L} = \mathcal{L}_{heatmap} + \lambda \cdot \mathcal{L}_{focal}
\end{equation}

\noindent
where $\mathcal{L}_{heatmap}$ is the per-pixel BCE loss of the gaze target heatmap.

Through extensive testing, we demonstrate focal loss acts as a powerful auxiliary loss to enable smoother convergence of the in/out classification. The focal loss regulates the latent representation of the decoder's feature map which complements the pixel-independent heatmap prediction to improve overall performance.

\subsection{Augmentation}

Prior approaches~\cite{attendedtarget2020,gazelle2024,vitgaze2024} indicate three commonly used input augmentations, including 1) random image cropping which randomly zooms in an image whilst preserving head bounding boxes and gaze target within the frame, 2) horizontal flipping, and 3) random jittering of head bounding boxes. Note that data for gaze target tasks are often acquired from well-lit and high-fidelity visible data such as COCO dataset and movies. We hypothesize more imaging augmentation techniques are needed to enhance model robustness and generalization. We introduce an additional suite of photometric augmentations, namely, 4) colour jittering, 5) random grayscaling, 6) auto-contrasting, and 7) random sharpness adjustment. Experiment results validate the significance of these augmentation steps.

\section{Evaluation}
\label{sec:eval}

\begin{table*}[t]
\footnotesize
\centering
\caption[comparison]{Methods comparison on GazeFollow and VideoAttentionTarget (VAT) datasets.}
\label{table:comparison}
\begin{tabular}{ccccccc} \toprule
      \multirow{2}{*}{Method}  & \multirow{2}{*}{Learnable Params} & \multicolumn{2}{c}{GazeFollow} & \multicolumn{3}{c}{VideoAttentionTarget} \\
     &  & AUC$\uparrow$ & Mean L2$\downarrow$  &  AUC$\uparrow$ & Mean L2$\downarrow$ & AP$_{in/out}\uparrow$ \\ \midrule
     Human Expert & - & 0.924 & 0.096  & 0.921 & 0.051 & 0.925 \\ \midrule
     Recasens et al.~\cite{gazefollow2015} & 50M & 0.878 & 0.19 & - & - & - \\
     Chong et al.~\cite{chong2018} & 51M & 0.896 & 0.187  & 0.833 & 0.171 & 0.712 \\
     Chong et al.~\cite{attendedtarget2020} & 61M & 0.921 & 0.137  & 0.86 & 0.134 & 0.853 \\
     Fang et al.~\cite{Fang_dual_attention2021} & 68M & 0.922 & 0.124  & 0.905 & 0.108 & 0.896 \\
     Hu et al.~\cite{rgbd_gaze2022} & 61M & 0.923 & 0.128  & 0.88 & 0.118 & 0.881 \\
     Gupta et al.~\cite{gupta2022} & 35M & 0.943 & 0.114  & 0.914 & 0.11 & 0.879 \\
     Tafasca et al.~\cite{childplay2023} & 25M & 0.939 & 0.122  & 0.914 & 0.109 & 0.834 \\
     ESCNet~\cite{escnet2022} & 29M & 0.928 & 0.122  & 0.885 & 0.12 & 0.869 \\
    Gaze-LLE$_{ViT-B}$~\cite{gazelle2024}  & 2.8M & 0.956 & 0.104  & 0.933 & 0.107 & 0.897 \\
    Gaze-LLE$_{ViT-L}$~\cite{gazelle2024}  & 2.9M & \textbf{0.958} & \textbf{0.099}  & 0.937 & 0.103 & 0.903 \\ \midrule
    {GazeMoE} & 3.4M & \textbf{0.959} & 0.101  & \textbf{0.939} & \textbf{0.097} & \textbf{0.917} \\ \bottomrule
\end{tabular}
\end{table*}

\begin{table*}[t]
\footnotesize
\centering
\caption[childplay]{Evaluation results on ChildPlay dataset. $\dagger$We test released Gaze-LLE models on the test set that the authors of ChildPlay dataset suggested.}
\label{table:childplay}
\begin{tabular}{cccc} \toprule
      \multirow{2}{*}{Method}  & \multicolumn{3}{c}{ChildPlay}  \\
      &  AUC$\uparrow$ & Mean L2$\downarrow$ & AP$_{in/out}\uparrow$\\ \midrule
      Gupta et al.~\cite{gupta2022} & 0.919 & 0.113 & 0.983 \\
    Tafasca et al.~\cite{childplay2023}  & 0.935 & 0.107 & 0.986 \\ 
    Gaze-LLE$_{ViT-B}^{\dagger}$~\cite{gazelle2024}  & 0.939 & 0.123 & 0.992 \\ 
    Gaze-LLE$_{ViT-L}^{\dagger}$~\cite{gazelle2024}  & 0.942 & 0.111 & 0.993 \\ \midrule
    GazeMoE & \textbf{0.945}  & \textbf{0.106}  & \textbf{0.994} \\ \bottomrule
\end{tabular}
\end{table*}

We conduct experiments on GazeFollow, VideoAttentionTarget, ChildPlay, GazeFollow360, EYEDIAP datasets, and ablation studies to unveil GazeMoE's optimum setting and runtime performance. We evaluate GazeMoE and comparative methods using the following metrics: area under the curve (AUC) of the probability heatmap, mean error distance (Mean L2) of peak gaze location, average precision (AP) of in-frame and out-of-frame classification, and a proprietary spherical distance for 360-degree images of GazeFollow360~\cite{gazefollowing360_2021}. All experimental results are derived from an average of five repeated training campaigns of different seeds.

The input image resolution is set to $448\times448$. The output heatmap is of $64\times 64$ in consistence with previous works. We use three transformer blocks in the decoder with 8 self-attention heads. The models are first pre-trained on GazeFollow for 15 epochs with batch size 60 and a $1e-3$ learning rate. Then, the models are fine-tuned on respective datasets with batch size 36, $1e-3$ as the learning rate for the in/out classification head and $1e-5$ as the learning rate for other layers. We utilize \textit{Adam} optimizer, a $0.1$ dropout, and a cosine learning rate scheduler.

\subsection{Comparative Study}

\noindent
\textbf{GazeFollow}. We pre-train the GazeMoE on GazeFollow~\cite{gazefollow2015} first, a large-scale dataset containing over 110K training samples from diverse visual scenes. The recommended full stack of augmentations are applied in conjunction with per pixel BCE loss. As is seen from Table~\ref{table:comparison}, the GazeMoE demonstrates competitive performance with state-of-the-art AUC. Note the advances of GazeMoE is not significant compared to Gaze-LLE models. This is primarily because GazeFollow is known to collect data samples from well-lit, high-fidelity imaging datasets like COCO and Flickr. Selective feature routing and sharing is therefore not remarkable on this benchmark.

\vspace{0.15cm}
\noindent
\textbf{VideoAttentionTarget (VAT)}. Chong et al.~\cite{attendedtarget2020} expanded the challenge by curating a new gaze target benchmark - VideoAttentionTarget (VAT) - to simultaneously classify in-frame and out-of-frame targets. We conduct supervised fine-tuning on VAT for 10 epochs combining heatmap BCE loss and the auxiliary focal loss in Eqn.~\ref{eqn:loss}. We set $\lambda=1$. As shown in Table~\ref{table:comparison}, GazeMoE outperforms prior methods demonstrating state-of-the-art performance with a significant margin over prior methods.

\vspace{0.15cm}
\noindent
\textbf{ChildPlay}. Tafasca et al.~\cite{childplay2023} recently created the ChildPlay dataset containing over 250K instances of child-to-child and child-to-adult gaze interactions from diverse Youtube sources. Previous datasets mainly focus on adult gazes, whereas, children's gaze can be tricky to understand due to the particular stage of their cognitive development. Besides in-frame and out-of-frame, we note that ChildPlay also annotates other gaze classes, such as \textit{gaze-shift}, \textit{eyes-closed}, and \textit{uncertain}, accounting for $10\%$ of total data. We exclude other classes before testing GazeMoE in comparison to existing methods in Table~\ref{table:childplay}. We set $\lambda=0.1$ and train for 3 epochs. Again, our method represents state-of-the-art solution to ChildPlay.

\begin{table*}[!t]
\footnotesize
\centering
\caption[gf360]{Evaluation results on GazeFollow360 dataset.}
\label{table:gf360}
\begin{tabular}{ccc} \toprule
      \multirow{2}{*}{Method}  & \multicolumn{2}{c}{GazeFollow360}  \\
      &  Spherical Dist$\downarrow$ & AUC$\uparrow$ \\ \midrule
      Human Expert & 0.2531 & 0.935 \\ \midrule
      Random & 1.5357 & 0.5056 \\
    Lian et al.~\cite{Lian2018}  & 1.254 & 0.6057 \\ 
    SALICON~\cite{SALICON2015} & 1.094 & 0.8002 \\
    Chong et al.~\cite{chong2018} & 0.9183 & 0.7765 \\
    Ruiz et al.~\cite{head_keyless2018} & 0.7612 & 0.7188 \\
    Zhang et al.~\cite{appearance_gaze2017} & 0.7366 & 0.7213 \\
    Li et al.~\cite{gazefollowing360_2021} & \textbf{0.6067} & 0.8104 \\ 
    Gaze-LLE$_{ViT-L}$~\cite{gazelle2024}  & 0.759 & 0.9197 \\ \midrule
    GazeMoE &  0.771  & \textbf{0.9232} \\ \bottomrule
\end{tabular}
\end{table*}

\begin{table*}[!t]
\centering
\footnotesize
\caption[eyediap]{Evaluation results on EYEDIAP dataset.}
\label{table:eyediap}
\begin{tabular}{lccc} \toprule
    \multirow{2}{*}{Method} & \multicolumn{3}{c}{EYEDIAP}  \\
     & AUC$\uparrow$ & Mean L2$\downarrow$ & AP$_{in/out}\uparrow$ \\ \midrule
    Gaze-LLE$_{ViT-B}$  & 0.617 & 0.411 & 0.725\\
    Gaze-LLE$_{ViT-L}$  & 0.596 & 0.421 & 0.614 \\
    Gaze-LLE$_{ViT-B}$ + head prompt  & 0.597 & 0.423 & \textbf{0.73} \\
    Gaze-LLE$_{ViT-L}$ + head prompt  & 0.593 & 0.431 & 0.662 \\ \midrule
    GazeMoE & \textbf{0.618} & \textbf{0.312} & 0.702  \\
    \bottomrule
\end{tabular}
\end{table*}

\begin{table*}[!t]
\footnotesize
\centering
\caption[ablation3]{Ablation study on different loss functions.}
\label{table:ablation3}
\begin{tabular}{ccccccc} \toprule
      \multirow{2}{*}{Loss}  & \multirow{2}{*}{Type} & \multicolumn{2}{c}{GazeFollow} & \multicolumn{3}{c}{VideoAttentionTarget} \\
     &  & AUC$\uparrow$ & Mean L2$\downarrow$  &  AUC$\uparrow$ & Mean L2$\downarrow$ & AP$_{in/out}\uparrow$ \\ \midrule
    \multirow{2}{*}{Heatmap Head} & BCELoss (GazeMoE) & \textbf{0.959} & \textbf{0.101}  & 0.939 & \textbf{0.097} & \textbf{0.917} \\
      & MSELoss + KL-D & 0.955 & 0.108  & 0.938 & 0.106 & 0.909 \\\midrule
    \multirow{2}{*}{In/out Head} & Focal Loss (GazeMoE) & - & - &  0.939 & 0.097 & \textbf{0.917} \\ 
     & BCELoss & - & -  & 0.938 & 0.098 & 0.912 \\ \bottomrule
\end{tabular}
\end{table*}

\begin{figure*}[t]
    \centering
    \includegraphics[width=0.87\linewidth]{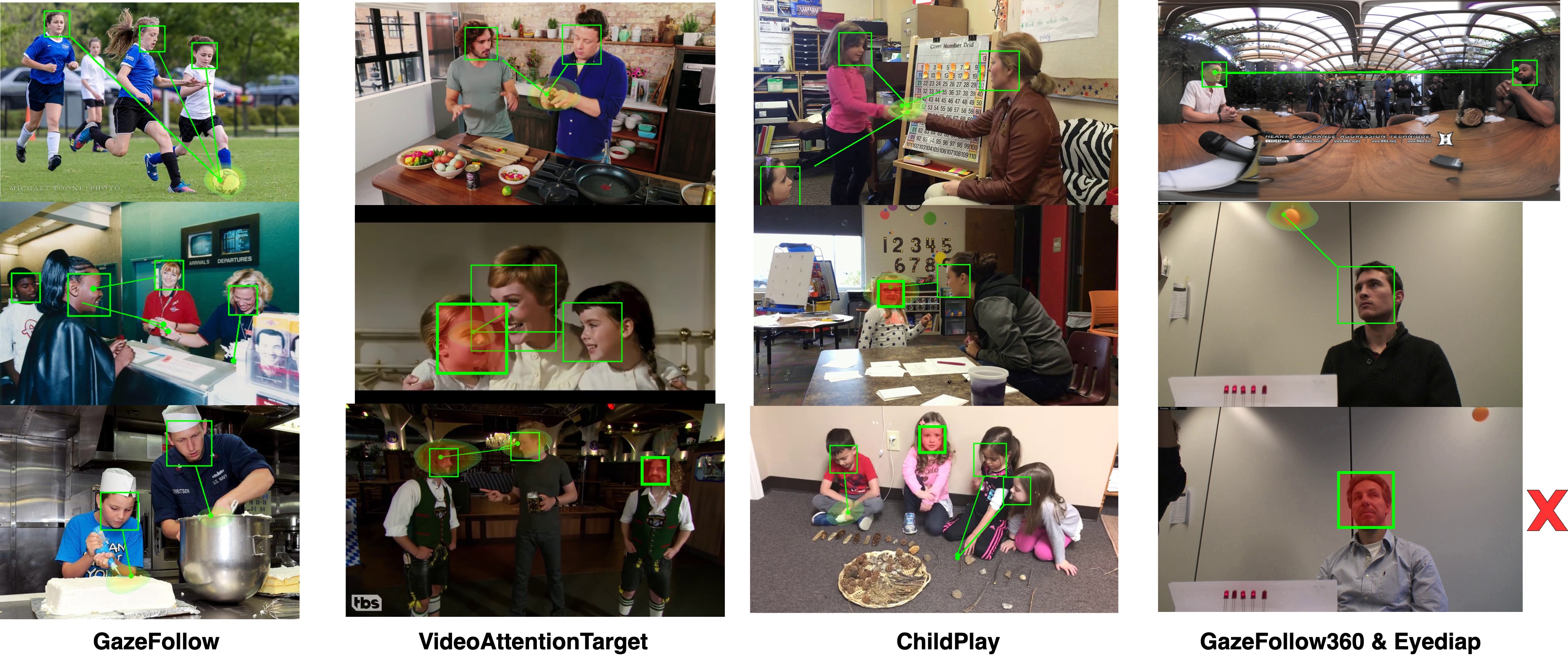}
    \caption{Qualitative results. We use \textit{dlib} face detector to automate head bounding boxes prompting as green frames.}
    \label{fig:qualitative}
\end{figure*}

\vspace{0.15cm}
\noindent
\textbf{GazeFollow360}. GazeFollow360 contains 360-degree images with omnidirectional field-of-views. These images undergo heavy distortion brought by sphere-to-plane projection raising difficulties for gaze target evaluation. The original work~\cite{gazefollowing360_2021} used sophisticated multiple pathways to diminish the assumption of light-of-sight in the frame. We follow the same split of 8,225 images for training, about 900 for validation, and 900 for testing. We conduct fine-tuning on GazeFollow360 with a batch size of 36 and a $1e-4$ learning rate. The models converge after 10 epochs. As shown in Table~\ref{table:gf360}, the evaluation results give a strong indication that GazeMoE adapts well to out-of-distribution data, surpassing previous methods on AUC and approaching human-level performance. This verifies that a pre-trained DINOv2 excels in gaze target estimation regardless of line-of-sight hypothesis.

\vspace{0.15cm}
\noindent
\textbf{EYEDIAP zero-shot inference}. The EYEDIAP~\cite{eyediap2014} dataset curates videos of human participants continuously gazing at a floating ping-pong ball in a laboratory. We evenly sample 50 frames from these videos and construct a dataset of 1,750 samples including $38.6\%$ out-of-frame images where the ping-pong ball is beyond camera's view. Because participants always fixated at the ball, it is unsuitable for supervised fine-tuning where models quickly overfit to attend to the ball. We conduct zero-shot testing as shown in Table~\ref{table:eyediap}. Again, GazeMoE delivers state-of-the-art performance. The MoE architecture allows the model to adaptively integrate gaze-related cues in challenging unseen scenarios.

\subsection{Ablation Study}
\label{sec:abla}

\textbf{Loss function}. We conduct extensive experiments to validate GazeMoE settings. We prove that BCE loss for $hm$ with an auxiliary focal loss for the in/out classification is optimal (Table~\ref{table:ablation3}), though MSE loss with a $0.05$ weighted KL-divergence (KL-D) loss can still lead to convergence.

\noindent
\textbf{MoE configuration}. We investigated the effects of MoE configuration including vanilla decoder without MoE (\textit{FFN only}), ratios between expert hidden size and transformer's $d_{model}$ (k$\,mlp$), and the number of shared experts (m$\,shared$) in Table~\ref{table:ablation1}. Performance boost brought by MoE module is apparent in comparison to vanilla feedforward decoders, GazeMoE$_{FFN\,only}$.

\noindent
\textbf{Augmentation}. We verify the efficacy of the proposed suite of augmentations as shown in Table~\ref{table:ablation2} and at the same time verify the robustness of the proposed GazeMoE. Random cropping in junction with photometric augmentations are essential to optimize the model though the MoE decoder already pushes performance to a high level without augmentations.

\begin{table}[!t]
\footnotesize
\centering
\begin{minipage}[t]{0.48\textwidth}
\centering
\caption[ablation1]{Evaluation on MoE size.}
\label{table:ablation1}
\begin{tabular}{ccc} \toprule
      \multirow{2}{*}{Model}  & \multicolumn{2}{c}{GazeFollow}  \\
      &  AUC$\uparrow$ & Mean L2$\downarrow$ \\ \midrule
      GazeMoE$_{FFN\,only}$ & 0.955 & 0.105 \\
      GazeMoE$_{1shared,\,2mlp}$ & 0.957 & \textbf{0.101} \\
      GazeMoE$_{1shared,\,4mlp}$ & 0.956 & 0.103 \\ 
      GazeMoE$_{2shared,\,1mlp}$ & 0.958 & \textbf{0.101} \\ \midrule
      GazeMoE$_{1shared,\,1mlp}$ &  \textbf{0.959}  & \textbf{0.101} \\ \bottomrule
\end{tabular}
\end{minipage}%
\hfill
\vspace{0.2cm}
\begin{minipage}[t]{0.48\textwidth}
\centering
\caption[ablation2]{Evaluation of augmentations.}
\label{table:ablation2}
\begin{tabular}{ccc} \toprule
      \multirow{2}{*}{Augmentations}  & \multicolumn{2}{c}{GazeFollow}  \\
      &  AUC$\uparrow$ & Mean L2$\downarrow$ \\ \midrule
      No Augmentation & 0.954 & 0.111 \\
      Cropping & 0.957 & \textbf{0.101} \\
      Photometric & 0.956 & 0.103 \\ 
      Cropping + Photometric &  \textbf{0.959}  & \textbf{0.101} \\ \bottomrule
\end{tabular}
\end{minipage}
\end{table}

\subsection{Qualitative Analysis}

We showcase the efficacy of GazeMoE across various datasets as can be seen from Figure~\ref{fig:qualitative}. GazeMoE works robustly in multi-person, diverse scenes including outdoor sports, TV advertisements, children learning environments, different public and private locations. It also performs well identifying in-frame vs. out-frame gazing (with a red colour bounding box over the face). The example where the algorithm has not performed well is EYEDIAP where there are extreme gaze angles plus it is zero-shot.

\subsection{Inference Evaluation and Implementation}

\begin{table}[!t]
\footnotesize
\centering
\caption[runtime]{Runtime performance.}
\label{table:runtime}
\begin{tabular}{ccc} \toprule
    Method & Latency (ms) &  Memory (MB) \\ \midrule
    GT360~\cite{gt360} & 70.3 & 689 \\
    Gaze-LLE$_{ViT-L}$  & 72.5 & 930 \\ \midrule
    GazeMoE &  74.2  & 984 \\ \bottomrule
\end{tabular}
\end{table}

We evaluate GazeMoE's runtime performance with an Intel Core i7 processor and an NVIDIA GeForce RTX 3080 GPU as shown in Table~\ref{table:runtime}. Per-sample latency of GazeMoE equals 74.2~ms, comparable to that of Gaze-LLE$_{ViT-L}$~\cite{gazelle2024} and GT360~\cite{gt360} which claimed to be an efficient solution.
This allows GazeMoE performing at 13 frames per second, which is sufficient for a number of applications such as human-robot interaction. GazeMoE's peak memory usage is 984~MB. Although it is greater than Gaze-LLE$_{ViT-L}$ and much more than GT360 at peak, GazeMoE realizes state-of-the-art performance making it an appealing solution in terms of precision and robustness.

GazeMoE is implemented using Ubuntu 22.04 vanilla with PyTorch 2.5.1 with NVIDIA CUDA support enabled. CUDA library required is v11.8 or higher. 

\section{Conclusion}
\label{sec:conclusion}


We propose a novel Mixture-of-Experts decoder for human gaze target estimation in imaging data. Based on this approach, we introduce GazeMoE, which leverages a unique combination of shared and routed experts for adaptive gaze feature learning under different scenarios. Our proposed model is the first to utilize a specialized Mixture-of-Experts structure in gaze target tasks. In all the datasets being evaluated - GazeFollow, VideoAttentionTarget, ChildPlay, GazeFollow360, EYEDIAP - our model outperforms existing methods to the best of our knowledge, which shows the potential to become a new benchmark for future gaze target research. The runtime performance of GazeMoE clearly sheds light on capabilities of deploying GazeMoE in robotic systems for real-time processing.

In future work, our research will focus on investigating and developing methods that exploit temporal information and contextual cues to handle extreme head poses and partially visible faces in real-world scenarios.


\section*{Acknowledgement}

Experiments were run on Aston Engineering and Physical Science Machine Learning Server, funded by the EPSRC Core Equipment Fund, Grant EP/V036106/1. The authors would like to thank to UKRI EPSRC EP/Y028813/1 ``National Edge AI Hub for Real Data: Edge Intelligence for Cyberdisturbances and Data Quality.'' The authors would like to acknowledge support by Villum Experiment project (00058627) and the Aston Research Flourish Together Support Fund.









\bibliographystyle{unsrt}
\bibliography{ref}

\end{document}